\documentclass[10pt, twocolumn, letterpaper]{article}

\usepackage{cvpr}              % To produce the CAMERA-READY version
\usepackage{graphicx}
\usepackage{amsmath}
\usepackage{amssymb}
\usepackage{booktabs}
\usepackage[accsupp]{axessibility} % Improves PDF readability for those with disabilities.
\newcommand{\daniel}[1]{\textcolor{black}{#1}}

\usepackage[pagebackref,breaklinks,colorlinks]{hyperref}
\usepackage{tabularx}

\usepackage[capitalize]{cleveref}
\crefname{section}{Sec.}{Secs.}
\Crefname{section}{Section}{Sections}
\Crefname{table}{Table}{Tables}
\crefname{table}{Tab.}{Tabs.}

\def\cvprPaperID{20} % *** Enter the CVPR Paper ID here
\def\confName{CVPR}
\def\confYear{2023}

\begin{document}

\title{DETR-based Layered Clothing Segmentation and Fine-Grained Attribute Recognition} 

\author{Hao Tian \qquad Yu Cao \qquad P.~Y.~Mok\thanks{Corresponding author}\\
The Hong Kong Polytechnic University, 
Kowloon, Hong Kong\\
{\tt\small {\{hao-henry.tian, yu-daniel.cao\}@connect.polyu.hk, tracy.mok@polyu.edu.hk}}
}

\maketitle

%%%%%%%%% ABSTRACT %%%%%%%%%
\begin{abstract}
    Clothing segmentation and fine-grained attribute recognition are challenging tasks at the crossing of computer vision and fashion, which segment the entire ensemble clothing instances as well as recognize detailed attributes of the clothing products from any input human images. Many new models have been developed for the tasks in recent years, nevertheless the segmentation accuracy is less than satisfactory in case of layered clothing or fashion products in different scales. In this paper, a new DEtection TRansformer (DETR) based method is proposed to segment and recognize fine-grained attributes of ensemble clothing instances with high accuracy. In this model, we propose a \textbf{multi-layered attention module} by aggregating features of different scales, determining the various scale components of a single instance, and merging them together. We train our model on the Fashionpedia dataset and demonstrate our method surpasses SOTA models in tasks of layered clothing segmentation and fine-grained attribute recognition.
\end{abstract}

%%%%%%%%% INTRO %%%%%%%%%

\section{Introduction}
\label{sec:intro}

Clothing segmentation and attribute recognition are the fundamental pre-tasks in many fashion applications such as outfit matching, fashion recommendation, and virtual try-on. With a growing interest in fashion related AI research, many researchers devoted themselves to this field and presented excellent work \cite{survey}. However, existing methods still face drawbacks when performing layered clothing segmentation and fine-grained attribute recognition tasks.
Ge \emph{et al.} \cite{deepfashion2} proposed a Match R-CNN to integrate clothing detection, landmark regression, segmentation, and retrieval into such a multi-task learning framework trained on 
%their own 
DeepFashion2 dataset. 
However, %most of these models 
their method is deficient in the case of clothing vague or layered occlusion and unable to handle multiple tasks harmoniously and simultaneously.
Jia \emph{et al.} \cite{fashionpedia} proposed the Attribute-Mask R-CNN to jointly perform instance segmentation and localized attribute recognition on Fashionpedia dataset with the whole ensemble of clothing instances.
Nevertheless, the gap between clothing segmentation and attribute recognition still exists, as well as the incomplete and inferior results for fine-grained attributes.
\daniel{To bridge the gap between instance segmentation and attribute recognition, Xu \emph{et al.} \cite{fashionformer} presented Fashionformer by building a DETR-based \cite{detr}  framework trained on Fashionpedia dataset~\cite{fashionpedia} with a Multi-Layer Rendering module for the attribute stream to explore more fine-grained features.} However, their method is not sensitive to clothing or accessories with scale differences and results in missing or incomplete clothing segmentation.

Considering that the tasks of segmentation and attribute recognition complement to one another, we design a model based on DETR that is equipped with a \textbf{\textit{multi-layered attention module}} for the segmentation stream in the decoder design. Our method achieves higher average precision and better visual results in clothing segmentation, and addresses to the issue of incomplete mask in case of clothing with different scales.
Our main \emph{contributions} are summarized as follows: (1) We build a DETR-based model that can perform well on layered clothing segmentation and fine-grained attribute recognition task; (2) We proposed a new multi-layered attention module in our decoder \daniel{by aggregating features of fashion items with different scales;} and (3) Both qualitative and quantitative comparative analyses show that our method outperforms other SOTA methods in terms of layered clothing segmentation and fine-grained attribute recognition performance.

%-------------------------------------------------------------------------
\begin{figure*}
  \centering
    \includegraphics[width=.95\textwidth]{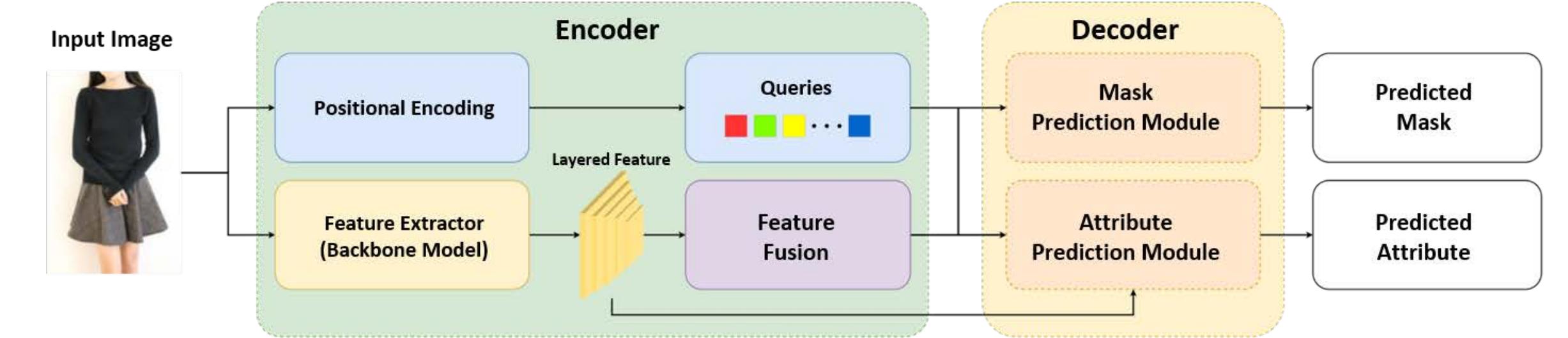} 
    \caption{The overall architecture of the DETR-based method with detailed decoder structure illustrated in Fig.~\ref{fig:block}}
    \vspace{-5px}
    \label{fig:arch}
\end{figure*}
%%%%%%%%% RELATED WORK %%%%%%%%%
\section{Related Work}

{\bf Layered Clothing Segmentation} 
Many researchers have attempted various approaches to the task of multi-layer occluded clothing segmentation and attribute recognition. 
Zheng \emph{et al.} \cite{modanet} achieved clothing segmentation and attribute recognition with Faster RCNN, SSD, and YOLO on their ModaNet dataset.
Jia \emph{et al.} \cite{fashionpedia} proposed a dataset named Fashionpedia and designed a novel Attribute-Mask R-CNN model to realize the multi-label attribute prediction. 

{\bf Fashion Datasets} \daniel{In the field of fashion segmentation and attribute recognition, there are currently mainly 4 publicly available datasets, DeepFashion \cite{liu2016deepfashion}, ModaNet \cite{modanet}, Deepfashion2 \cite{deepfashion2} and Fashionpedia \cite{fashionpedia}.} Fashionpedia is a step toward mapping out the visual aspects of the fashion world, which consists of an ontology built by fashion experts and a dataset with everyday and celebrity event fashion images annotated. 
Unlike the DeepFashion series, images in Fashionpedia mainly focused on the whole ensemble of clothing. ModaNet \cite{modanet} focused on outwear and accessories, nevertheless ModaNet server is no longer accessible. We therefore trained the model on the Fashionpedia dataset and conducted comparative experiment. %made some inferences. 

{\bf Transformer-based Detection and Segmentation}
Carion \emph{et al.} \cite{detr} first presented an end-to-end transformer structure to handle object detection task, which also introduced the concept of object query which is a type of input to the transformer decoder. Shi \emph{et al.}~\cite{shi2022transformer} used internal attention and external attention for multi-level context mining. \emph{Inspired by these works, a DETR-based transformer architecture is proposed to unify and simplify fashion tasks. The main contribution of this work lies in the decoder design the transformer structure, covering both the segmentation prediction and attribute prediction modules.}
%-------------------------------------------------------------------------
\begin{figure}[b]
  \centering
    \includegraphics[width=.95\linewidth]{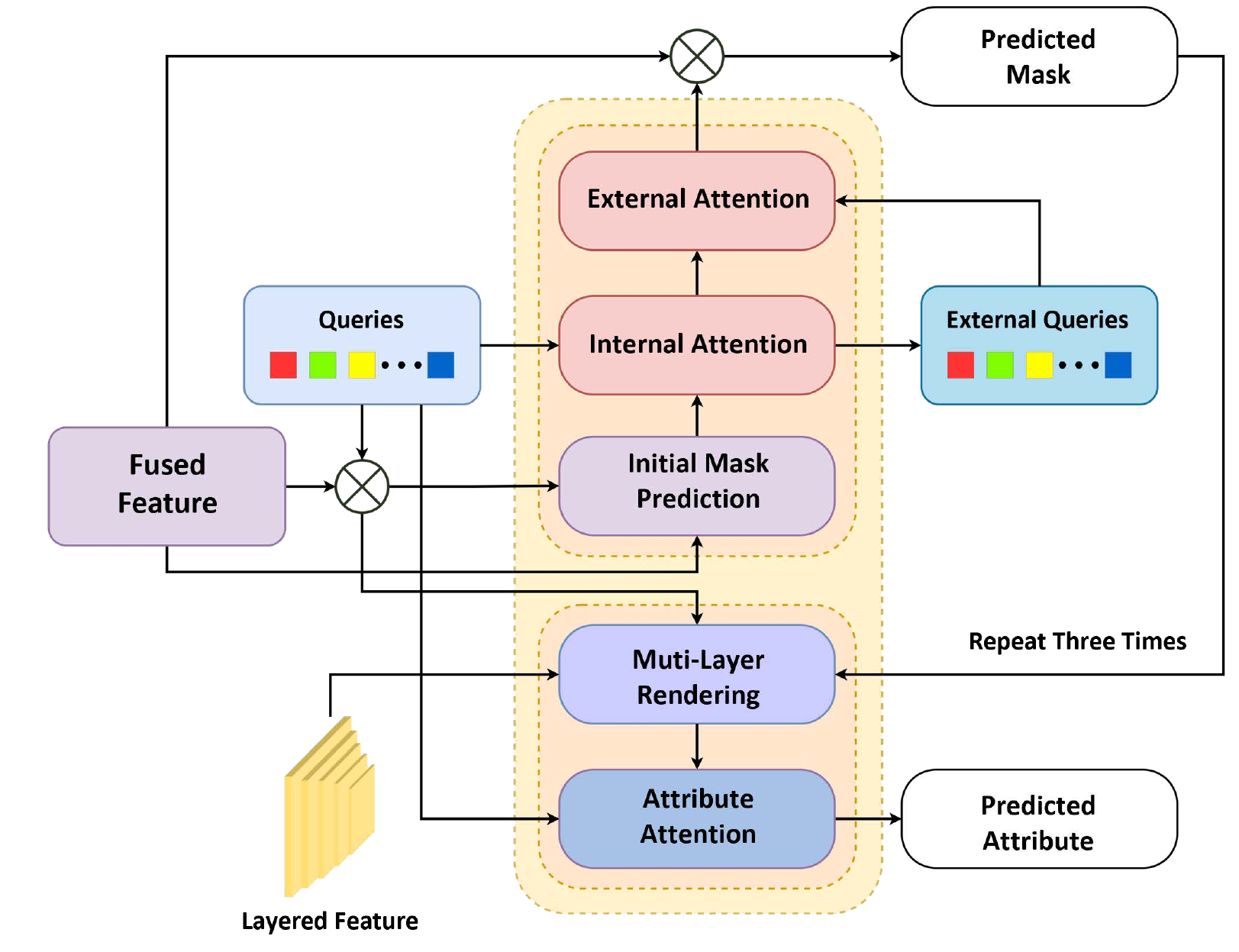} 

   \caption{The multi-layered attention module structure in the decoder.}
   \label{fig:block}
\end{figure}
%%%%%%%%% METHOD %%%%%%%%%
\section{Method}

%-------------------------------------------------------------------------
\subsection{Overall Architecture}

Fig.~\ref{fig:arch} shows the overall architecture of our method, depicting the encoder and decoder parts. In the encoder, we first obtain layered features from the input image by a backbone network with the feature pyramid structure, then fuse the layered feature map into a fused feature and a positional encoder generates positional embeddings at the same time, which we called $Query$, simplified as $Q$. The layered feature, fused feature, and queries are then sent to the decoder. The decoder consists of a mask prediction module and an attribute prediction module. Similar to~\cite{fashionformer}, the mask prediction module and the attribute prediction module of our model work in a cascaded mode. Instead of generating mask by mask grouping and query learning~\cite{fashionformer}, our method takes fused features and queries as inputs for the mask prediction module, taking into account of both global and local features to improve the mask prediction accuracy, especially for the layered clothing in different scales. %\tocheck{Tracy: It it not clear here. Need to rewrite based on how~\cite{fashionformer} did in this.}

%-------------------------------------------------------------------------
\subsection{Encoder}

Firstly, we send each input image into the feature extractor to obtain the layered feature map $F_i$, where $i$ $\in \{1,2,3,4\}$ are indexes of different scales. The feature extractor is a backbone network with a feature pyramid structure, such as Mask R-CNN and Swin Transformer. Then we use a concatenate function to sum up the layered features into a fused feature map $F_f$.
A $1\times1$ convolution layer is settled to generate a $d$$\times$$H$$\times$$W$ feature map, which is the query $Q$. We also obtain the weights of the instance masks in this step, which are also equal to queries weights. The queries are significant input for two prediction modules in the decoder.

%-------------------------------------------------------------------------

\begin{figure*}
  \centering
  \begin{subfigure}{0.19\linewidth}
  \includegraphics[width=.95\linewidth]{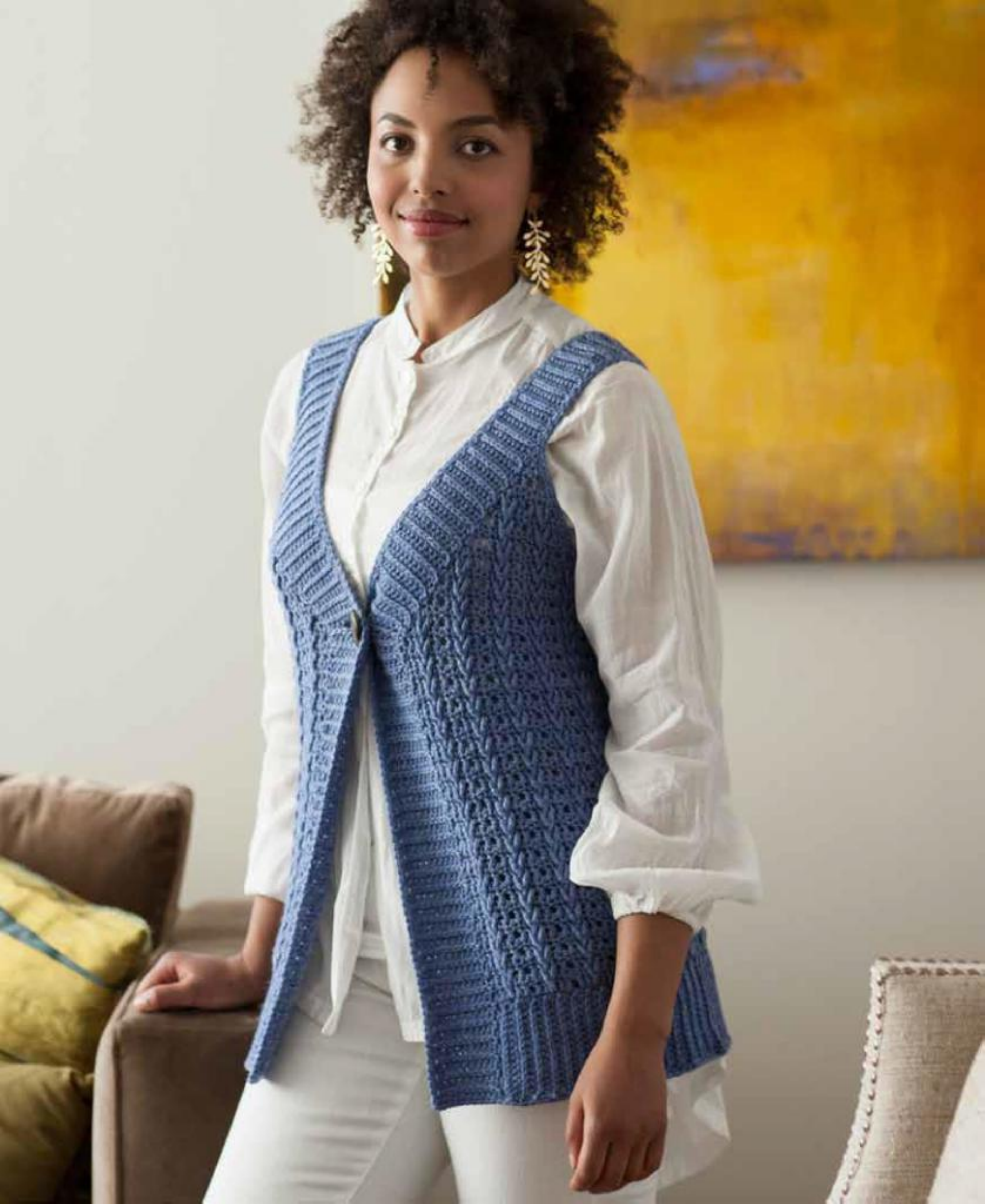} 
    \caption{Original Image}
    \label{fig:OI}
  \end{subfigure}
  \hfill
  \begin{subfigure}{0.19\linewidth}
    \includegraphics[width=.95\linewidth]{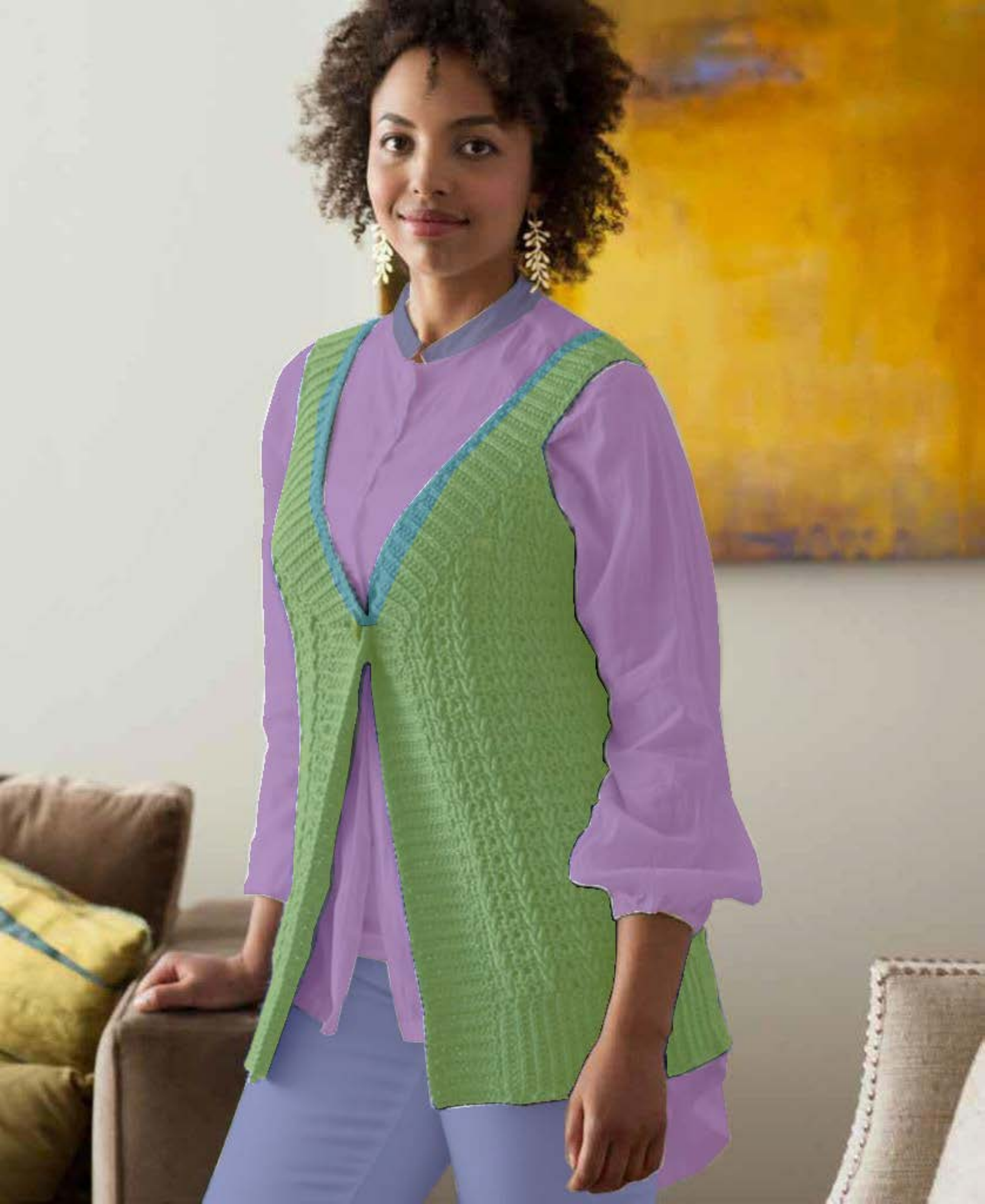} 
    \caption{Groud Truth}
    \label{fig:GT}
  \end{subfigure}
  \hfill
  \begin{subfigure}{0.19\linewidth}
    \includegraphics[width=.95\linewidth]{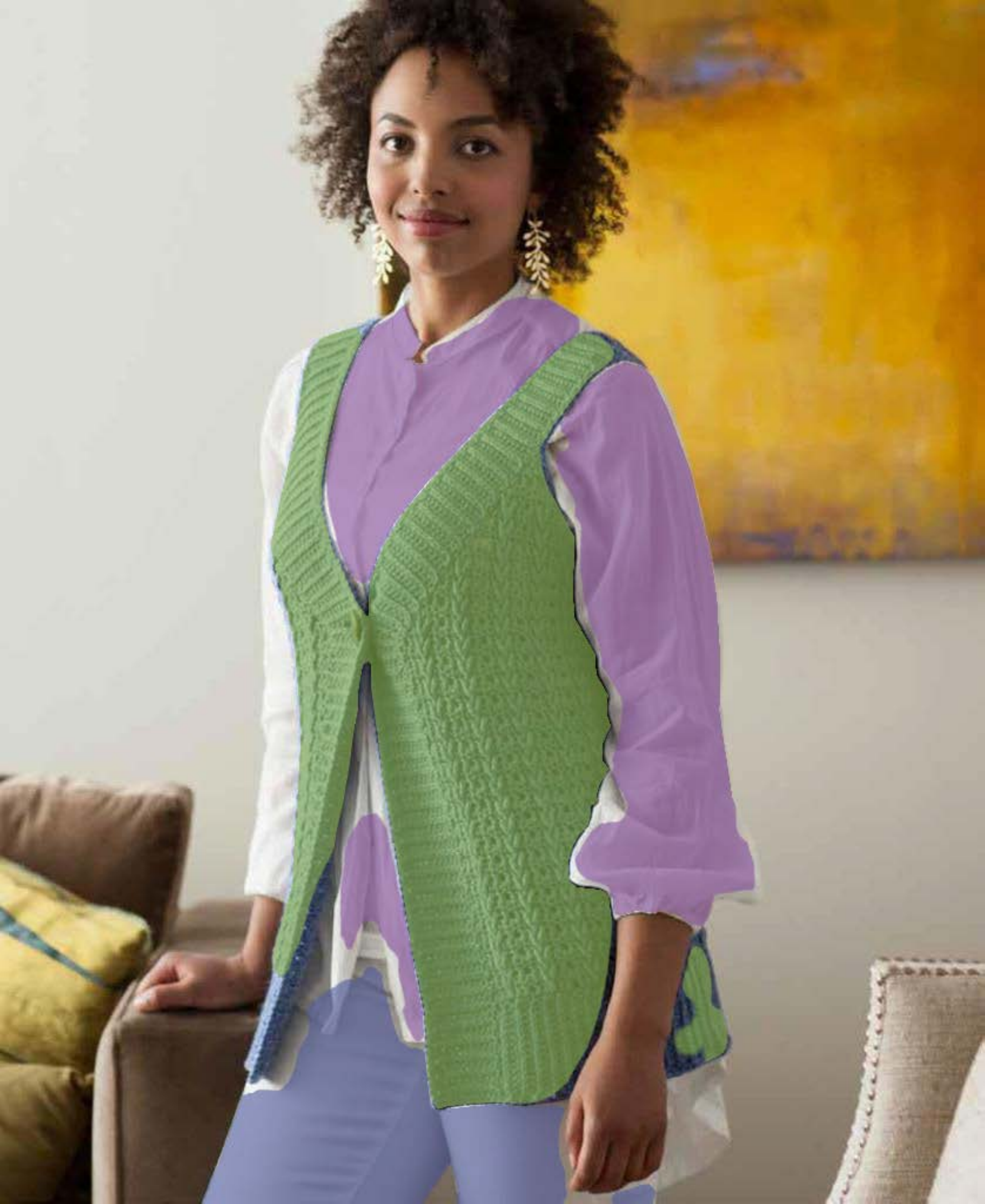} 
    \caption{Attribute-Mask RCNN~\cite{fashionpedia}}
    \label{fig:AM}
  \end{subfigure}
  \hfill
  \begin{subfigure}{0.19\linewidth}
    \includegraphics[width=.95\linewidth]{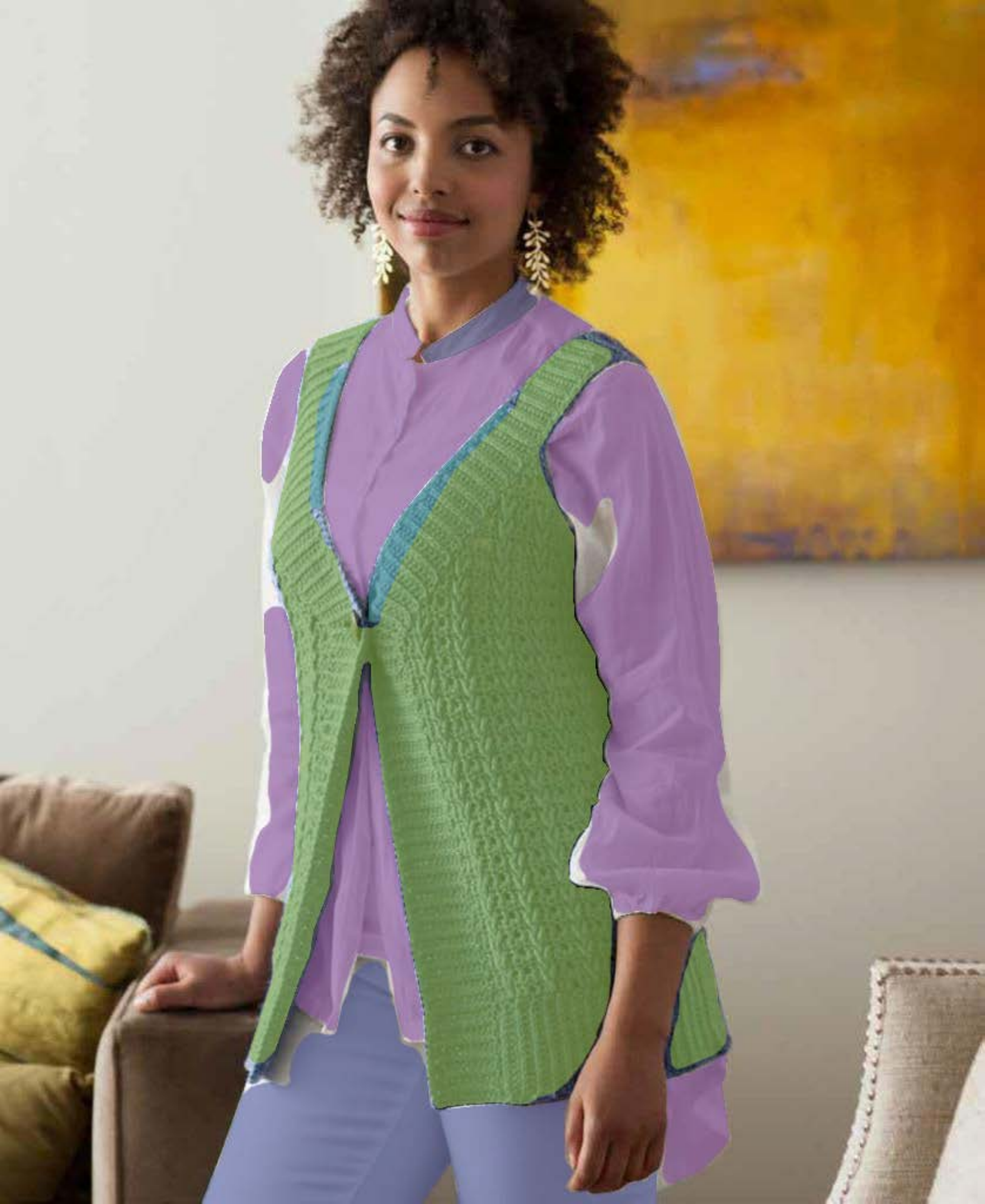} 
    \caption{FashionFormer~\cite{fashionformer}}
    \label{fig:FF}
  \end{subfigure}
  \hfill
  \begin{subfigure}{0.19\linewidth}
    \includegraphics[width=.95\linewidth]{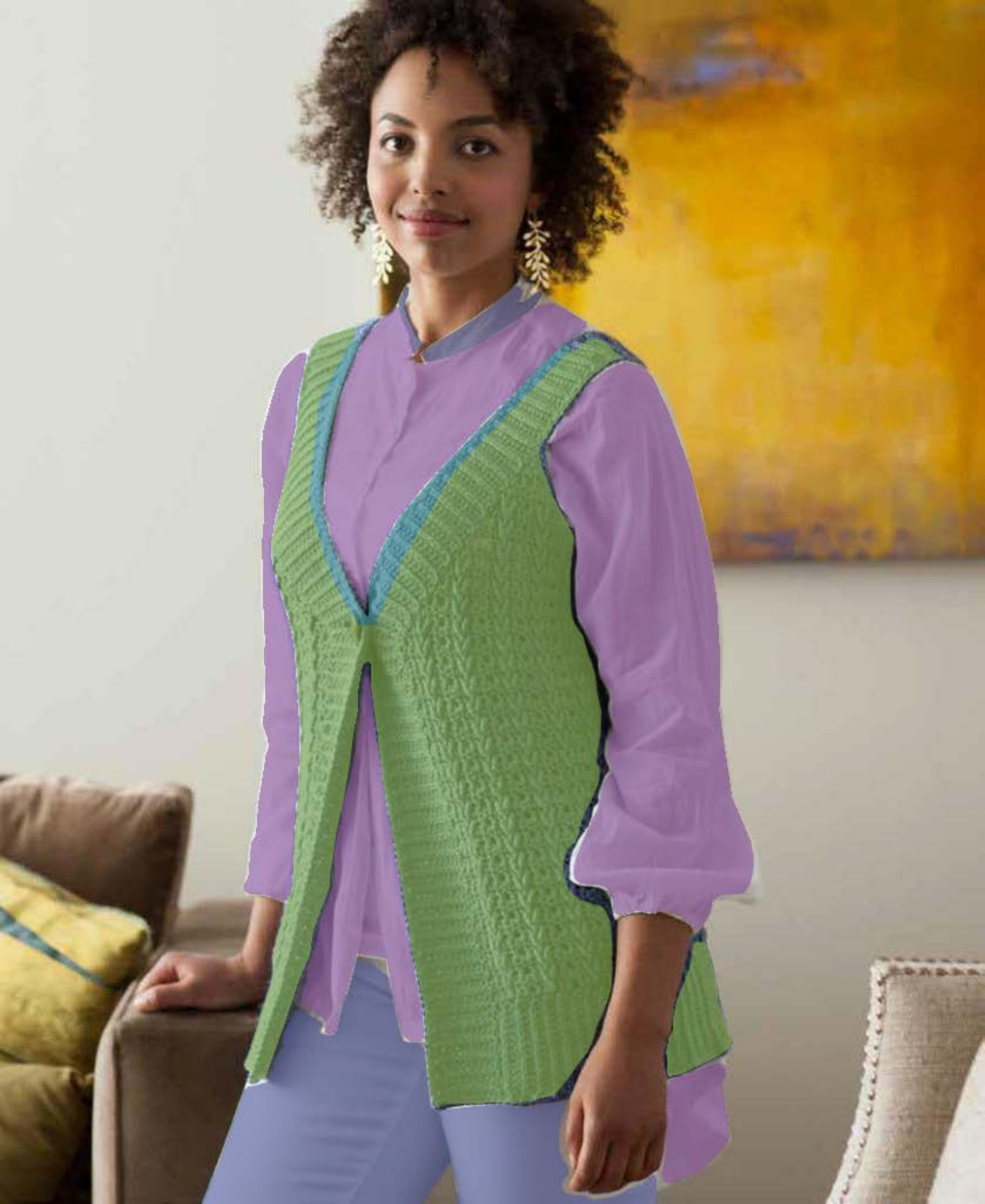} 
    \caption{\textbf{Ours}}
    \label{fig:OURS}
  \end{subfigure}
\vspace{-5px}
  %\caption{Here are the experiment results, and the details show that our method achieves the best clothing instance segmentation results than other SOTA methods.}
  \caption{Experimental result comparison with other SOTA methods.}
  \label{fig:res}
\end{figure*}

%-------------------------------------------------------------------------
\begin{figure*}
  \centering
  \begin{subfigure}{0.37\linewidth}
  \includegraphics[width=.95\linewidth]{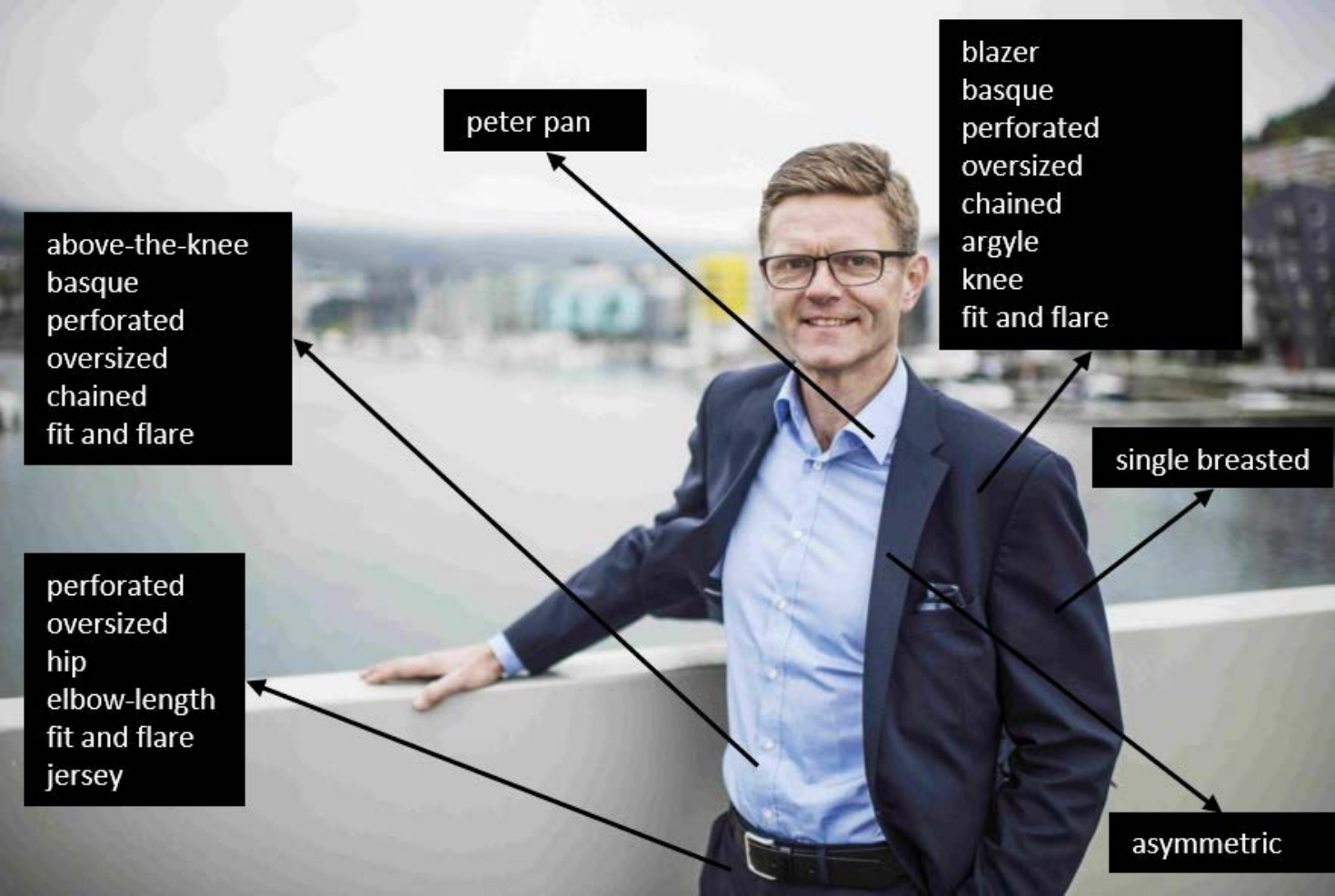} 
    \caption{Input Image and Ground Truth Attributes}
    \label{fig:OI}
  \end{subfigure}
  \begin{subfigure}{0.37\linewidth}
    \includegraphics[width=.95\linewidth]{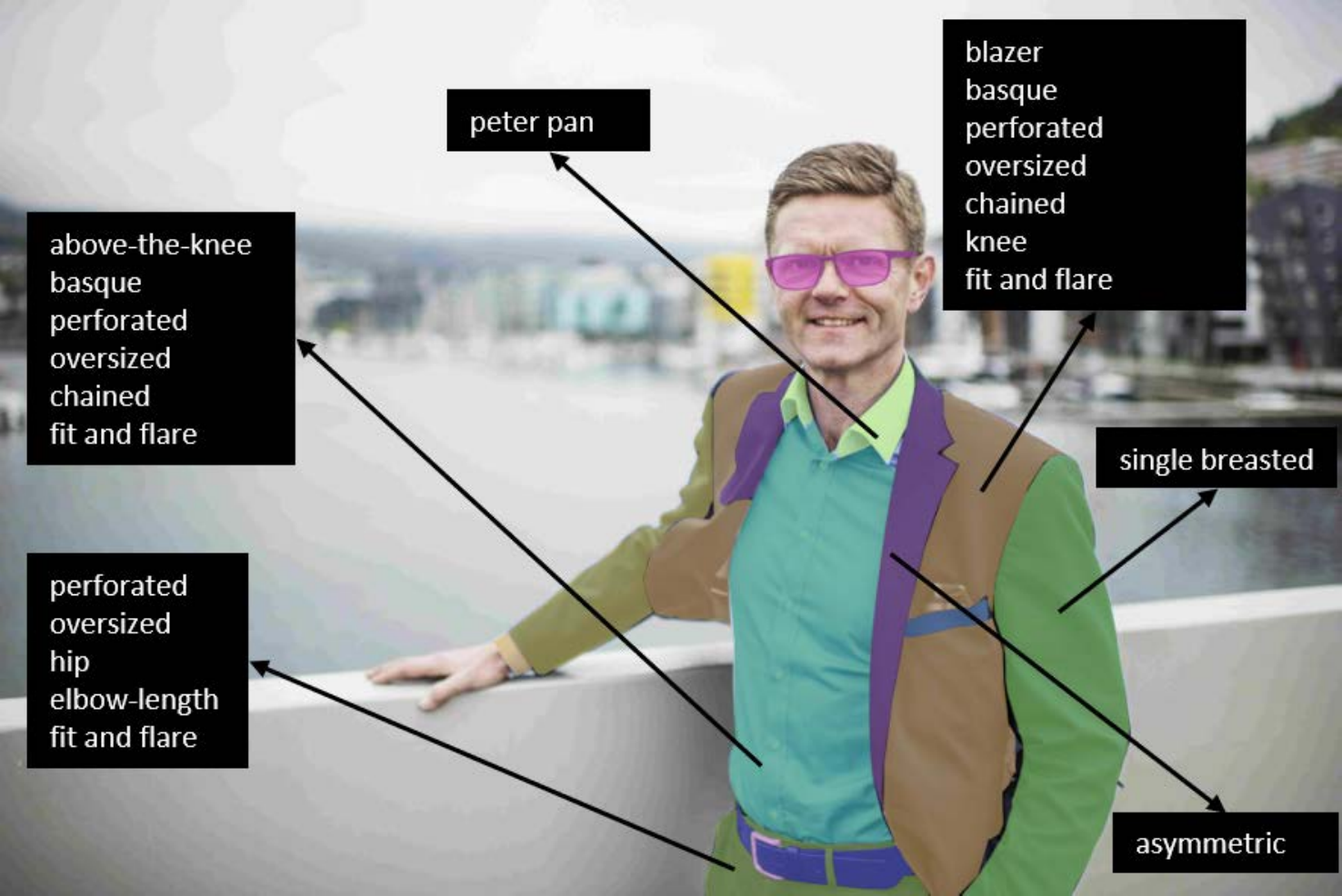} 
    \caption{Our Masks and Attributes Results}
    \label{fig:GT}
  \end{subfigure}
%  \hfill
  %\caption{Visual results with attributes.}
 \vspace{-5px}
 \caption{An example result of attribute recognition.}
  \label{fig:att}
\end{figure*}
%-------------------------------------------------------------------------
\subsection{Decoder}

We build the decoder with a mask prediction module and an attribute prediction module, as shown in Fig.~\ref{fig:block}.
For the mask prediction module, we first adopt a multi-layered attention module with layered features and fused features to obtain the initial mask prediction. We compute the mask prediction by the self-attention in the internal attention and external attention module as the equation below:
\begin{eqnarray}
P_{mask} = SoftMax \left( {Q K^{T} \over\sqrt{C}} \right )  F
\end{eqnarray}
where $K$ is the key positional embeddings, $C$ is the vector dimensions in internal attention and external attention respectively. 
The fused feature and Queries perform Kronecker product operation to generate the prediction of the initial mask.
For internal attention, $Q$ are the queries, $K$ and $v$ come from the initial mask prediction, then the result of this attention can generate a new prediction and a series of new queries which we called $External Queries$. Finally, external attention uses external queries as $Q$ and predictions generated by internal attention to predict next-generation mask prediction, which perform Kronecker product operation with fused feature to generate the final predicted mask.

In the attribute prediction module, we followed the Multi-Layer Rendering (MLR) in \cite{fashionformer}, so our attribute recognition results are fine-grained as well. The multi-level features are computed as:
\begin{eqnarray}
F_{i,atr}^{j} = \sum_{u}^{W_{i}}\sum_{v}^{H_{i}}P^{j-1}\left(u, v, i \right )\cdot F_{i}\left(u, v \right ) \cdot Q_{i}^{j}\left(u, v \right )
%, i \in \left \{1, 2, 3, 4\right \} 
\end{eqnarray}
where $j$ is the interaction number, $j$ $\in \{1,2,3\}$, $W_{i}$ and $H_{i}$ are the height and width of the corresponding features $F_{i}$, $u$ and $v$ are the spatial index of features.

%-------------------------------------------------------------------------
\subsection{Loss Function}
By simultaneously considering mask, class, and attribute results, we use bipartite matching as the cost. 
We use the focal loss \cite{lin2017focal} in mask prediction and classification. For mask prediction, we use dice loss \cite{milletari2016v}.
As is the case with multi-label classification tasks, our model's predictions are supervised in a one-hot format. For attributes, we follow the default design of Attribute-Mask R-CNN \cite{fashionpedia}.
It is an attribute loss with many binary labels. Here is how the entire loss might be expressed:
\begin{eqnarray}
    L = \sum_{j=1}^{3} {\lambda_{cls} L_{cls}^{(j)} + \lambda_{mask} L_{mask}^{(j)} + \lambda_{attribute} L_{attribute}^{(j)}}
\end{eqnarray}
where ${j}$ means the training stage index and here all $\lambda$ are empirically set to 1.

%-------------------------------------------------------------------------
\begin{table*}[t]
  \centering
  \begin{tabular}{@{}lcccccc@{}}
%\begin{tabularx}{0.8\textwidth}{@{}lcccccc@{}}
   \toprule
    Method & Backbone & Schedule & Flops(B) & Params(M) & $AP_{IoU}^{mask}$ & $AP_{IoU+F_{1}}^{mask}$\\
    \midrule
    Attribute-Mask R-CNN~\cite{fashionpedia} & \begin{tabular}[c]{@{}l@{}}R50-FPN\\R101-FPN\\Swin-B\end{tabular} & 3x  & \begin{tabular}[c]{@{}l@{}}296.7\\374.3\\508.3\end{tabular} & \begin{tabular}[c]{@{}l@{}}46.4\\65.4\\107.3\end{tabular} & \begin{tabular}[c]{@{}l@{}}39.2\\40.7\\47.5\end{tabular} & \begin{tabular}[c]{@{}l@{}}29.5\\31.4\\40.6\end{tabular} \\
    \midrule
    Fashionformer~\cite{fashionformer} & \begin{tabular}[c]{@{}l@{}}R50-FPN\\R101-FPN\\Swin-B\end{tabular} & 3x  & \begin{tabular}[c]{@{}l@{}}198.0\\275.7\\442.5\end{tabular} & \begin{tabular}[c]{@{}l@{}}37.7\\56.6\\100.6\end{tabular} & \begin{tabular}[c]{@{}l@{}}42.5\\45.6\\49.5\end{tabular} & \begin{tabular}[c]{@{}l@{}}39.4\\42.8\\46.5\end{tabular} \\
    \midrule
    Ours & \begin{tabular}[c]{@{}l@{}}R50-FPN\\R101-FPN\\Swin-B\end{tabular} & 3x & \begin{tabular}[c]{@{}l@{}}196.4\\270.2\\423.8\end{tabular} & \begin{tabular}[c]{@{}l@{}}36.5\\53.5\\94.7\end{tabular} & \begin{tabular}[c]{@{}l@{}}45.4\\48.2\\52.3\end{tabular} & \begin{tabular}[c]{@{}l@{}}39.8\\43.4\\46.9\end{tabular} \\
    \bottomrule
  \end{tabular}
%\end{tabularx}
  \caption{Results on Fashionpedia.}
  \label{tab:res}
\end{table*}

%%%%%%%%% EXPERIMENTS %%%%%%%%%
\section{Experiments}
\label{sec:formatting}

\subsection{Dataset, Metrics and Implementation Details}
We conduct experiments on the Fashionpedia \cite{fashionpedia} dataset.
The Fashionpedia dataset includes segmentation masks and attribute labels together, so it is appropriate for our model's training and inference. 

For clothing instance segmentation, we adopt the mean average precision $mAP_{IoU}^{mask}$ from the COCO setting. In order to evaluate clothing segmentation and attribute recognition together, we adopt $mAP_{IoU+F1}^{mask}$ from \cite{fashionformer}, which is a joint metric combining $mAP$ and $F1_{score}$.
The GFlops are obtained with 3 × 1020 × 1020 inputs following \cite{fashionpedia}.

We implement our method using the PyTorch framework with MMDetection toolbox \cite{mmdetection}, and the model is trained on 2 NVIDIA RTX 3090 GPUs with the batch size of 16. One standard training schedule 1x is 5625 iterations, and we set the training epoch number as 3x. The learning rate is set by \cite{goyal2017lr}. and both the generator and the discriminator are alternately updated in every iteration. We use large scale jittering \cite{ghiasi2021lsj} that resizes an image to a random ratio between [0.5, 2.0] of the target input image size. 

%-------------------------------------------------------------------------
\subsection{Qualitative Evaluation}

Fig.~\ref{fig:res} compares visual results generated by our method and other SOTA models on R101-FPN backbone. 
Almost every attribute of the input image as well as flawless instance segmentation masks can be predicted by our model.
Our model achieves the best segmentation results, especially for fashion items with large scale differences. Moreover, our method has better clothing segmentation and attribute results than other SOTA methods. 
As can be seen in Fig.~\ref{fig:att}, some tiny scale fashion items can be completely detected and segmented, and some small parts belonging to one clothing instance but scattered can also be detected and segmented into the correct instance. Furthermore, the attributes of the recognized instances are mostly complete and correct. It proves the effectiveness of the proposed \textbf{\textit{multi-layer attention module}} in our decoder design.

%-------------------------------------------------------------------------
\subsection{Quantitative Evaluation}
Table~\ref{tab:res} compares our method with Fashionformer \cite{fashionformer} and Attribute-Mask R-CNN \cite{fashionpedia} in different settings. By using R50-FPN backbone, our method outperforms Fashionformer and Attribute-Mask R-CNN, respectively, in terms of $mAP_{IoU}^{mask}$ by 2.9\% and 6.2\%; and in terms of $mAP_{IoU+F1}^{mask}$ by 0.4\% and 10.3\%. By using R101-FPN backbone, our method outperforms in terms of $mAP_{IoU}^{mask}$ by 2.6\% and 7.5\%, and in terms of $mAP_{IoU+F1}^{mask}$ by 0.6\% and 12.0\%. 
To compare with SpineNet \cite{du2020spinenet} with more training iterations, Xu \emph{et al.} \cite{fashionformer} adopted Swin-B as the backbone and retrained their model. Comparatively, we also retrained on Swin-B backbone, our method still outperforms Fashionformer by 2.8\% in $mAP_{IoU}^{mask}$ and 0.4\% in $mAP_{IoU+F1}^{mask}$, and is better than Attribute-Mask R-CNN by 4.8\% in $mAP_{IoU}^{mask}$ and 6.3\% in $mAP_{IoU+F1}^{mask}$.
This indicates the effectiveness of our proposed new \textbf{\textit{multi-layer attention module}} in our decoder by aggregating features of fashion items with different scales.

%%%%%%%%% CONCLUSION %%%%%%%%%
\section{Conclusion}

In this paper, we design a new DETR-based model for the task of layered clothing segmentation and fine-grained attribute recognition. By refining the decoder with a specially designed \textbf{\textit{multi-layered attention module}}, we introduce an effective solution to the task, especially for cases when fashion items are of different scales. Extensive experiments on the Fashionpedia dataset have demonstrated that our method achieves competitive performance comparing with other state-of-the-art models. 

\section*{Acknowledgements}
The work is supported in part by a grant from the Research Grants Council of the Hong Kong Special Administrative Region, China (Grant Number 152112/19E) and by the Innovation and Technology Commission of Hong Kong, under grant ITP/028/21TP.
%%%%%%%%% REFERENCES %%%%%%%%%
{\small
\bibliographystyle{ieee_fullname}
\bibliography{egbib}
}

\end{document}